\documentclass[letterpaper]{article} 
\usepackage{aaai2026}  
\usepackage{times}  
\usepackage{helvet}  
\usepackage{courier}  
\usepackage[hyphens]{url}  
\usepackage{graphicx} 
\usepackage{natbib}  
\usepackage{caption} 
\usepackage{algorithm}
\usepackage{algorithmic}

\usepackage{newfloat}
\usepackage{listings}
\DeclareCaptionStyle{ruled}{labelfont=normalfont,labelsep=colon,strut=off} 
\floatstyle{ruled}
\newfloat{listing}{tb}{lst}{}
\floatname{listing}{Listing}
\title{Speech-Aware Long Context Pruning and Integration for Contextualized Automatic Speech Recognition}
\author{
    Yiming Rong\textsuperscript{\rm 1}\textsuperscript{\rm 2}\equalcontrib,
    Yixin Zhang\textsuperscript{\rm 1}\textsuperscript{\rm 2}\equalcontrib,
    Ziyi Wang\textsuperscript{\rm 1}\textsuperscript{\rm 2}\equalcontrib,
    Deyang Jiang\textsuperscript{\rm 1}\textsuperscript{\rm 2},
    Yunlong Zhao\textsuperscript{\rm 1}\textsuperscript{\rm 2},
    Haoran Wu\textsuperscript{\rm 1}\thanks{Corresponding author.},
    Shiyu Zhou\textsuperscript{\rm 1}\footnotemark[2],
    Bo Xu\textsuperscript{\rm 1}\textsuperscript{\rm 2}\footnotemark[2]
}
\affiliations{
    \textsuperscript{\rm 1}The Key Laboratory of Cognition and Decision Intelligence for Complex Systems, Institute of Automation, Chinese Academy of Sciences, China\\
    \textsuperscript{\rm 2}School of Artificial Intelligence, University of Chinese Academy of Sciences, China\\
    \{rongyiming2022, zhangyixin2024, wuhaoran2018, shiyu.zhou, xubo\}@ia.ac.cn

}

\usepackage{bibentry}

\usepackage{xcolor}
\usepackage{amsmath}
\usepackage{amssymb}
\usepackage{graphicx}
\usepackage{multirow}
\usepackage{tabularx}
\usepackage{array}
\usepackage{booktabs}
\usepackage{pifont}
\usepackage{arydshln}
\usepackage{makecell}
\newcommand{\SAPsq}{SAP\textsuperscript{2}}

\begin{document}

\maketitle

\begin{abstract}
Automatic speech recognition (ASR) systems have achieved remarkable performance in common conditions but often struggle to leverage long-context information in contextualized scenarios that require domain-specific knowledge, such as conference presentations. This challenge arises primarily due to constrained model context windows and the sparsity of relevant information within extensive contextual noise. To solve this, we propose the \SAPsq ~method, a novel framework that dynamically prunes and integrates relevant contextual keywords in two stages. Specifically, each stage leverages our proposed Speech-Driven Attention-based Pooling mechanism, enabling efficient compression of context embeddings while preserving speech-salient information. Experimental results demonstrate state-of-the-art performance of \SAPsq~ on the SlideSpeech and LibriSpeech datasets, achieving word error rates (WER) of 7.71\% and 1.12\%, respectively. On SlideSpeech, our method notably reduces biased keyword error rates (B-WER) by 41.1\% compared to non-contextual baselines. \SAPsq~ also exhibits robust scalability, consistently maintaining performance under extensive contextual input conditions on both datasets.
\end{abstract}

\begin{links}
    \link{Code}{https://github.com/jymh/SAP2-ASR}
\end{links}

\section{Introduction}

Automatic speech recognition (ASR) systems have achieved remarkable performance in common scenarios, approaching human-level accuracy in routine speech recognition tasks~\cite{radford2023robust,gao2022paraformer,xu2025fireredasr}. However, current challenges no longer center around baseline transcription accuracy but instead focus on effectively recognizing speech in complex, context-rich scenarios~\cite{chang2021context, ni2024vilas, wang2024cieasr, sudo2024contextualized}. For example, in conference videos, speakers commonly utilize visual aids, particularly slides containing large amounts of text~\cite{wang2024slidespeech, wang2024slideavsr}. Such OCR-derived contexts usually include important entities and domain-specific terminology. In Figure~\ref{fig:case study}, despite that keywords like "glaucoma" constitute only a small portion of the speech content, they are crucial for comprehending domain-specific information. However, processing extensive textual inputs introduces considerable computational overhead and the risk of incorporating irrelevant information when the textual contexts span hundreds of thousands of words.

Existing contextualized ASR methods mainly follow two paradigms: (1) deep biasing E2E models that apply attention mechanisms to influence the decoding of ASR model~\cite{pundak2018deep, chang2021context, han2021cif, han2022improving, huang2023contextualized, sudo2024contextualized, yu2024lcb}, and (2) SpeechLLM models that treat contextual keywords as prompts~\cite{lakomkin2024end, yang2024mala, bai2024seed}. Deep biasing models face performance gaps compared with large-scale pre-trained models~\cite{lakomkin2024end} and exhibit architectural rigidity. SpeechLLM models are more flexible and robust when incorporating contextual keywords. However, when dealing with long context information, simply treating contextual keywords as prompts poses challenges to effective long-context processing and robust noise filtering.

Neuroscience study on speech recognition~\citep{sohoglu2012predictive} indicates that human perceptual clarity improves significantly when speech aligns with prior contextual knowledge, demonstrating the human brain's active integration of prior knowledge during speech recognition. Our motivation is to simulate this active integration process, prune and compress noisy contextual information to knowledge more aligned with speech contents. We propose \SAPsq ~(\textbf{S}peech-\textbf{A}ware Context \textbf{P}runing with \textbf{S}peech-Driven \textbf{A}ttention-based \textbf{P}ooling), a novel framework that actively filters long-form contextual keywords and integrates them into ASR systems without sacrificing efficiency. Our contributions are as follows:

\begin{figure*}[h]
    \centering
    \includegraphics[width=0.9\linewidth]{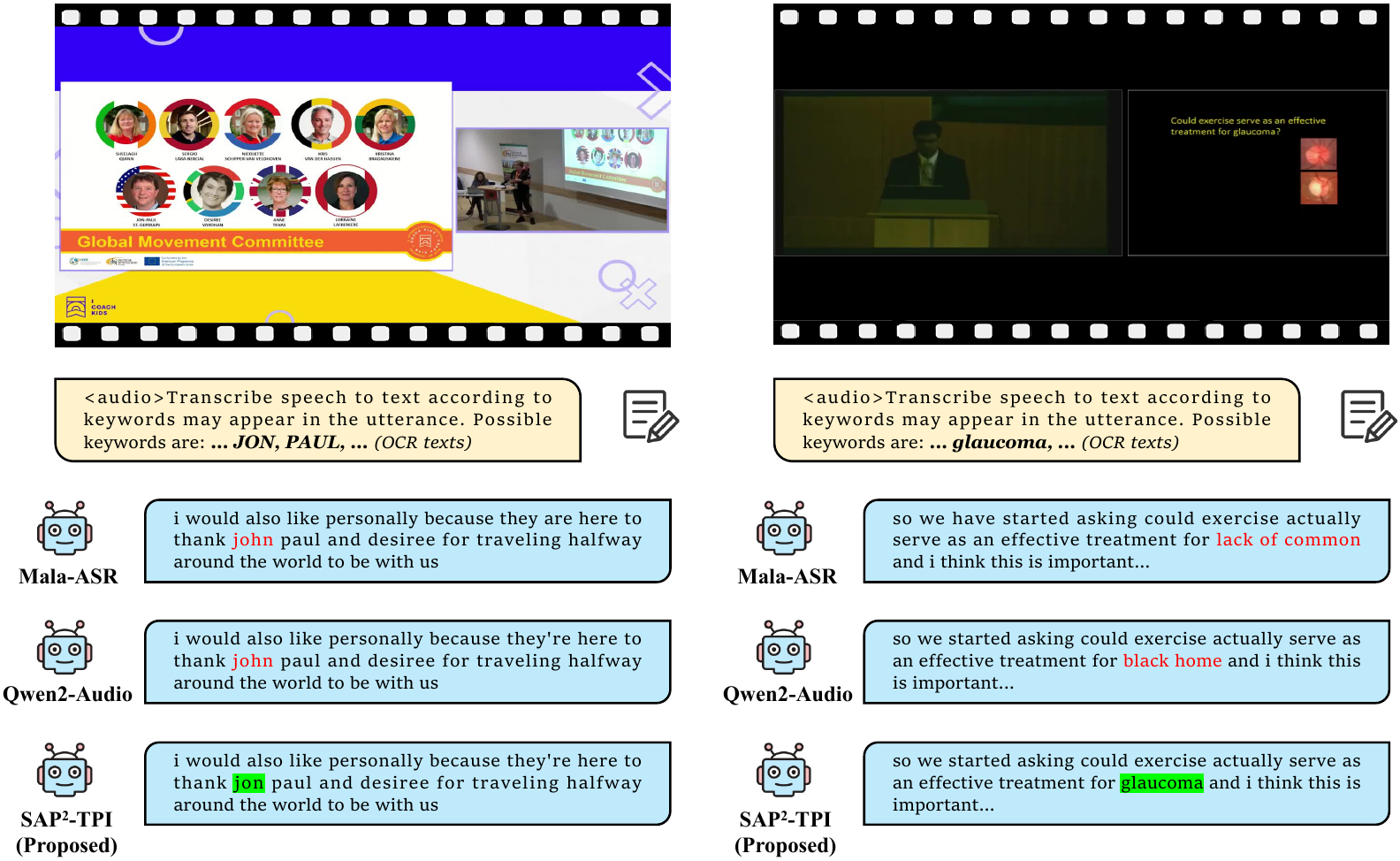}
    \caption{Comparisons of the previous SOTA MaLa-ASR, the baseline Qwen2-Audio-PC, and our \SAPsq-TPI on the SlideSpeech test set. For each sample, {\color{red}red} texts indicate recognition errors in proper nouns, while {\color{green}green}-highlighted texts showcase corrections made by \SAPsq-TPI. These two examples represent typical cases of proper noun recognition: the left demonstrates \SAPsq-TPI's accuracy in recognizing rare personal names, and the right highlights its capability in identifying medical terminology.}
    \label{fig:case study}
\end{figure*}

\begin{itemize}
    \item \textbf{Effective Long-Context Processing}: We introduce a novel Speech-Aware Context Pruning framework, instructing a SpeechLLM to dynamically prune OCR-derived textual contexts and retain only keywords directly relevant to the speech content.
    \item \textbf{Cross-Modal Context Compression}: We propose a Speech-Driven Attention-based Pooling mechanism that leverages cross-modal alignment between speech and text. This pooling strategy compresses extensive textual inputs into concise, speech-relevant context embeddings.
    \item  \textbf{State-of-the-Art Performance and Robustness} Experimental results on SlideSpeech and LibriSpeech validate the superiority of our proposed \SAPsq~ method, achieving a WER of 7.71\% and 1.12\%, respectively. Particularly, our approach achieves a notable 41.1\% relative improvement in biased keyword recognition over non-contextual baselines on SlideSpeech. Furthermore, \SAPsq~ demonstrates robust scalability, maintaining consistent performance under extensive contextual input conditions on both datasets.
\end{itemize}

\begin{figure*}[h]
  \centering
  \includegraphics[width=\linewidth]{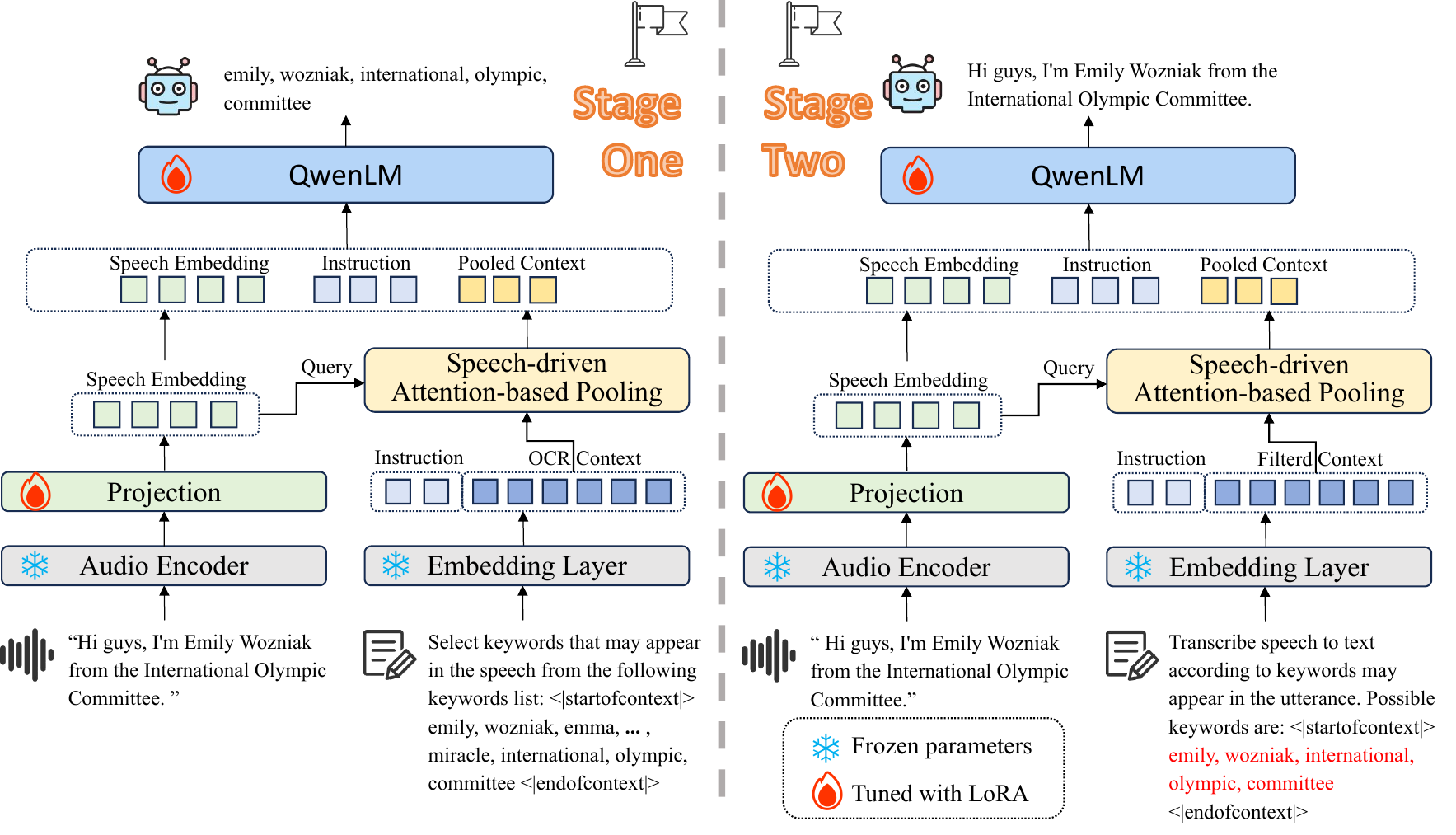}
  \caption{The overall architecture of proposed \SAPsq-TPI framework: In stage one, long contextual keywords are pruned based on speech to reduce irrelevant information. Pruned contextual keywords are inputs of stage two, used for contextualized speech recognition. Speech-driven attention-based pooling is utilized in both stages to compress context embeddings.}
  \label{model architecture}
\end{figure*}

\section{Method}

\subsection{Problem Formulation}

Given a sequence of acoustic features $\boldsymbol{X} = [\boldsymbol{x}_1, \ldots, \boldsymbol{x}_T]$ of length $T$ extracted by an audio encoder, the objective of speech recognition is to maximize the probability of output text label $\boldsymbol{Y} = [\boldsymbol{y}_1, \ldots, \boldsymbol{y}_L]$ of length $L$. In contextualized ASR scenarios, additional context information $\boldsymbol{Z} = [\boldsymbol{z}_1, \ldots, \boldsymbol{z}_C]$ is provided in the dataset $\mathcal{D} = \left\{\boldsymbol{X}^{(i)}, \boldsymbol{Y}^{(i)}, \boldsymbol{Z}^{(i)}\right\}_{i=1}^N$. The probability of an auto-regressive ASR model decoding the text label $\boldsymbol{Y}$ is formulated as:
\begin{equation} \label{eq: auto-regressive}
    p_{\theta}(\boldsymbol{Y}|\boldsymbol{X}, \boldsymbol{Z})=\prod_{j=1}^L p_{\theta}(\boldsymbol{y}_j|\boldsymbol{X}, \boldsymbol{Z}, \boldsymbol{Y}_{<j}),
\end{equation}
where $\theta$ is the parameter of the ASR model. For many deep biasing ASR models, $\boldsymbol{Z}$ is fed into a biasing decoder through cross-attention layers. In large pre-trained ASR models like Whisper or SpeechLLM models like Qwen2-Audio, $\boldsymbol{Z}$ is concatenated with acoustic features $\boldsymbol{X}$. We unify the presentation, ignore the implementation details and treat $\boldsymbol{Z}$ as a conditional variable. Thus, the overall training objective is:
\begin{equation}
    \max_{\theta} \mathbb{E}_{(\boldsymbol{X}, \boldsymbol{Y}, \boldsymbol{Z}) \sim \mathcal{D}}[ p_\theta(\boldsymbol{Y} | \boldsymbol{X}, \boldsymbol{Z}) ].
\end{equation}

\subsection{Speech-Aware Context Pruning}

For contextual keywords $\boldsymbol{Z} = [\boldsymbol{z}_1, \ldots, \boldsymbol{z}_C]$, when $C$ is a very large number, typically ten times larger than $L$, there might be few keywords truly relevant to the speech. This is a common case in conference recording videos, where slides related to a 10-second utterance contain thousands of words of background information. Suppose only $c$ core keywords $\tilde{\boldsymbol{Z}} = [\tilde{\boldsymbol{z}}_1, \ldots, \tilde{\boldsymbol{z}}_c]$ will benefit the recognition of $\boldsymbol{X}$, which means
\begin{equation} \label{eq:pruning decomposition}
p_{\theta}(\boldsymbol{Y}|\boldsymbol{X}, \boldsymbol{Z}) = \int p_{\theta}(\boldsymbol{Y}|\boldsymbol{X}, \tilde{\boldsymbol{Z}}) p(\tilde{\boldsymbol{Z}} | \boldsymbol{X}, \boldsymbol{Z}) d\tilde{\boldsymbol{Z}}.
\end{equation}

In practical settings, when text label $\boldsymbol{Y}$ and context information $\boldsymbol{Z}$ are given, the truly beneficial core keywords $\tilde{\boldsymbol{Z}} = \boldsymbol{Z} \cap \boldsymbol{Y}$ is the keywords that appear in the label, the original dataset can be augmented as $\mathcal{D} = \left\{ \left( \boldsymbol{X}^{(i)}, \boldsymbol{Y}^{(i)}, \boldsymbol{Z}^{(i)}, \tilde{\boldsymbol{Z}}^{(i)} \right) \right\}_{i=1}^N
$. Thus, we can suppose that $\tilde{\boldsymbol{Z}}$ is a deterministic variable, $p_{\theta}(\boldsymbol{Y}|\boldsymbol{X}, \boldsymbol{Z})$ is derived as:
\[
p_{\theta}(\boldsymbol{Y}|\boldsymbol{X}, \boldsymbol{Z}) = p_{\theta}(\boldsymbol{Y}|\boldsymbol{X}, \tilde{\boldsymbol{Z}}) p(\tilde{\boldsymbol{Z}} | \boldsymbol{X}, \boldsymbol{Z}).
\]

Suppose that the context pruning model 
$p(\tilde{\boldsymbol{Z}} | \boldsymbol{X}, \boldsymbol{Z})$ is parameterized by $\psi$, the overall objective of the system is:
\begin{equation}
    \max_{\theta, \psi} \mathbb{E}_{(\boldsymbol{X}, \boldsymbol{Y}, \boldsymbol{Z}, \tilde{\boldsymbol{Z}}) \sim \mathcal{D}}\left[ p_{\theta}(\boldsymbol{Y}|\boldsymbol{X}, \tilde{\boldsymbol{Z}}) p_{\psi}(\tilde{\boldsymbol{Z}} | \boldsymbol{X}, \boldsymbol{Z}) \right] .
\end{equation}

This derivation decomposes the original task into two stages. Prior work like \citet{jayanthi2023retrieve} explores this two-stage prune-and-integrate pipeline. However, the pruning model $p_{\psi}$ is mainly based on similarity-based algorithms, which makes it hard to generalize to unseeing textual contents and speech. In Figure~\ref{model architecture}, we utilize a SpeechLLM model as the context pruning model. It accepts the input of long contextual keywords and speech, and output core keywords relevant to the speech using speech-driven attention-based pooling. The core keywords generated in step one is concatenated with the speech to perform contextualized ASR. Regarding this \textbf{Two-stage Pruning-Integration (TPI)} manager, we denote the pipeline as \SAPsq-TPI. Unless specified otherwise, \SAPsq ~refers to the two-phase trained variant (\SAPsq-TPI). During the training time, ASR model $p_\theta$ and speech-driven context pruning model $p_\psi$ can be trained simultaneously, which means the training objective can be written as
\begin{equation} \label{overall objective} \small
    \max_{\theta} \mathbb{E}_{(\boldsymbol{X}, \boldsymbol{Y},  \tilde{\boldsymbol{Z}}) \sim \mathcal{D}}[ p_{\theta}(\boldsymbol{Y}|\boldsymbol{X}, \tilde{\boldsymbol{Z}})] \max_{\psi} \mathbb{E}_{(\boldsymbol{X}, \boldsymbol{Z}, \tilde{\boldsymbol{Z}}) \sim \mathcal{D}} [p_{\psi}(\tilde{\boldsymbol{Z}} | \boldsymbol{X}, \boldsymbol{Z})] .
\end{equation}

During the inference stage, $p_{\theta}(\boldsymbol{Y}|\boldsymbol{X}, \tilde{\boldsymbol{Z}})$ relies on $\tilde{\boldsymbol{Z}}$ produced by the context pruning model, so the speech recognition task is conducted after contextual keywords are pruned.

\subsection{Speech-driven Attention-based Pooling}

In SpeechLLM models, when the length of contextual keywords $C$ is too large or even exceeds the maximum context window of the LLM, the input inevitably suffers from content truncations or critical semantic loss. To solve this, we propose a Speech-driven Attention-based Pooling method that reduces the context length while preserving speech-relevant information. The idea is to weight contextual keywords with speech attention scores and conduct pooling on the weighted text features. We calculate the attention score between speech embedding $\boldsymbol{h}^x$ and the contextual keywords embedding $\boldsymbol{h}^z$ to determine the weight of tokens. Then we perform window-wise pooling to contract the embedding of keywords.
Please refer to the Appendix for deductions and detailed implementations.

This speech-driven attention-based pooling mechanism can be used in both the SpeechLLM ASR model and the context pruning model, combining the cross-attention module in deep biasing ASR models and simplicity of SpeechLLM ASR models.

\section{Experiments}

In this section, we conduct experiments to investigate the effectiveness of our framework.

\subsection{Experimental Setup}
\subsubsection{Dataset} 

We conduct experiments on the \textbf{SlideSpeech}~\citep{wang2024slidespeech} dataset and the \textbf{LibriSpeech} \cite{LibriSpeech} dataset. The contextual keywords of SlideSpeech are slide-derived OCR texts, and the contextual keywords of LibriSpeech is dynamically constructed following the approach in \citet{Le2021ContextualizedSE-DB}.

To investigate the impact of extended contextual information on speech recognition accuracy and to evaluate the effectiveness of our proposed long-context pruning and integration framework, we augment SlideSpeech OCR contexts by extracting semantic keywords from five consecutive slides. Specifically, using the Jaccard index to group slides into coherent clusters, combining multiple consecutive slides together. 

To assess how augmented contexts would influence ASR performance, we analyze two metrics:

\begin{itemize}
    \item \textit{\textbf{Keyword Coverage Rate}} (The ratio of core keywords in ASR texts to text length) increases from 4.01\% (single slide) to 6.51\% (five slides), confirming more contextual keywords that may aid ASR are included.
    \item \textit{\textbf{Information Rate}} (The ratio of core keywords to total keywords) decreases from 1.82\% (single slide) to 0.64\% (five slides), reflecting much higher noise levels in extended contexts. This demonstrates the critical need for robust noise-resistant methods to extract salient information from longer OCR contexts.
\end{itemize}

\begin{figure}[h]
  \centering
  \includegraphics[width=0.8\linewidth]{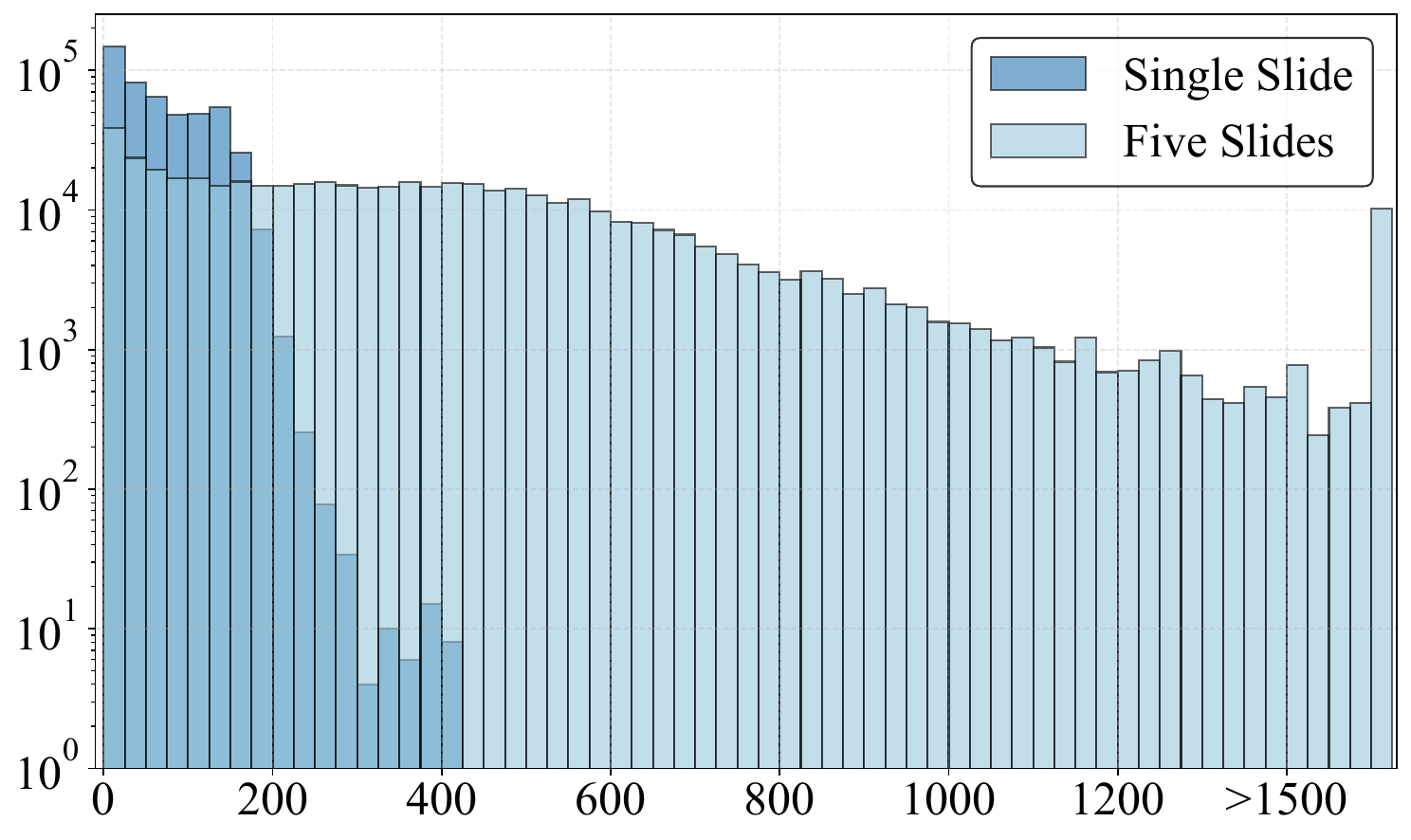}
  \caption{Distribution of tokenized keyword counts. (x-axis: number of tokens, linear scale; y-axis: instance count, log scale). Single-slide contexts exhibit mean/median token lengths of 64.57/53, while five-slide contexts reach 402.81/332.}
  \label{fig: keywords distribution}
\end{figure}

\subsubsection{Evaluation Metrics} 

We follow the evaluation metrics of previous works of contextualized ASR~\cite{huang2023contextualized, yu2024lcb, yang2024mala, wang2024slidespeech}. We evaluate the results using WER, biased word error rate (B-WER), unbiased word error rate (U-WER), and the Recall of words in both the biasing list and transcript. U-WER is calculated for words not in the contextual keywords list, which reflects the foundational ASR capabilities of the model. B-WER is computed for words in the contextual keywords list, which indicates the model's capability of accurately integrating core keywords. For contextualized ASR models, U-WER is mainly related to foundation ASR models, so it's more important to achieve improvements in B-WER while maintaining U-WER stability.


\begin{table*}[ht]
  \caption{Performance comparison of our \SAPsq-TPI and other models evaluated on SlideSpeech test dataset, trained on S95/L95.}
  \label{tab:main results}
  \centering
  \resizebox{0.8\linewidth}{!}{
  \begin{tabular}{llccccc}
    \toprule
    \multirow{2}{*}{\textbf{Train}} & 
    \multirow{2}{*}{\textbf{Model}} & 
     \multirow{2}{*}{\makecell{ \textbf{Contextual}\\ \textbf{Keywords}}}  & \multicolumn{4}{c}{\textbf{Test}} \\
    \cmidrule(lr){4-7} 
    & & &  \textbf{WER} & \textbf{U-WER} & \textbf{B-WER} & \textbf{Recall $\uparrow$}  \\
    \midrule
    \multirow{8}{*}{\makecell{S95\\(161h)}} & 
    SlidesSpeech~\citep{wang2024slidespeech} & \ding{55}&  21.22 & 20.83&26.60&73.51 \\
     & MaLa-ASR~\citep{yang2024mala} & \ding{55}&  11.80 & 11.71&13.52&86.71 \\ 
     & Qwen2-Audio~\citep{chu2024qwen2} & \ding{55} & 10.79 & 10.85 &7.92&90.33 \\ 
     \cmidrule(lr){2-7}
     & CPP~\citep{huang2023contextualized} & 1slide  & 20.95 & 20.73&24.05&76.10 \\
     & LCB-net~\citep{yu2024lcb} &1slide&  19.21 & 18.89&23.70&76.48 \\
     & MaLa-ASR~\citep{yang2024mala} &1slide&   11.26 & 11.52& 7.67 & 92.50 \\
     & \SAPsq-TPI (Proposed)  & 1slide  & 8.04  & \textbf{8.13} & 5.29 & 94.85 \\
     \cmidrule(lr){2-7}
     & \SAPsq-TPI (Proposed)  & 5slides&  \textbf{7.86} & 8.17 & \textbf{5.18} & \textbf{94.98} \\
    \midrule
    \multirow{8}{*}{\makecell{L95\\(473h)}} & SlidesSpeech~\citep{wang2024slidespeech} & \ding{55}&  12.89 & 12.90&12.70&87.43 \\
     & MaLa-ASR~\citep{yang2024mala}  & \ding{55}&   8.61& 8.72&7.34&92.84 \\
     & Qwen2-Audio~\citep{chu2024qwen2} & \ding{55}&  10.76 & 10.91 &7.71&91.42 \\ 
     \cmidrule(lr){2-7}
     & CPP~\citep{huang2023contextualized} & 1slide&   12.38 & 12.60&9.32&90.86\\ 
     & LCB-net~\citep{yu2024lcb} &1slide&  12.02 & 12.24&9.03&91.12 \\
     & MaLa-ASR~\citep{yang2024mala} & 1slide& 8.46 & 8.73& 4.89 & 95.31 \\
     & \SAPsq-TPI (Proposed)  & 1slide&  7,92 & 8.16 & 4.71 & 95.47 \\
     \cmidrule(lr){2-7}
     & \SAPsq-TPI (Proposed) & 5slides& \textbf{7.71} & \textbf{8.08} & \textbf{4.54} & \textbf{95.59} \\
    \bottomrule
  \end{tabular}
  }
\end{table*}

\subsection{Implementation Details}

We fine-tune Qwen2-Audio-7B-Instruct (hereafter referred to as Qwen2-Audio) for context pruning and speech recognition tasks. During fine-tuning, we employ LoRA adapters to update both the multimodal projector and LLM backbone while keeping the speech encoder frozen. Please refer to the Appendix for details of hyper-parameters.


Speech-driven attention-based pooling mechanism is applied in both stages. In each stage, we only compress contextual keywords, excluding instruction prompts. Without specific clarification, pooling window size is set to 2. Please refer to the Appendix for detailed instruction prompts.



\begin{table*}[h]
    \centering
    \caption{Performance (WER(U-WER/B-WER)) (\%) comparison of our \SAPsq-TPI method and different methods on the LibriSpeech test dataset with different biasing list length (N=100, 500, 1000).}
    \label{tab:LibriSpeech comparision}
    \begin{tabular}{l c c c c c c}
        \toprule
        \multirow{2}{*}{\textbf{Model}} 
        & \multicolumn{2}{c}{\textbf{N=100}} 
        & \multicolumn{2}{c}{\textbf{N=500}} 
        & \multicolumn{2}{c}{\textbf{N=1000}} \\
        \cmidrule(lr){2-3} \cmidrule(lr){4-5} \cmidrule(lr){6-7}
        & \textbf{test-clean} & \textbf{test-other} 
        & \textbf{test-clean} & \textbf{test-other} 
        & \textbf{test-clean} & \textbf{test-other} \\
        \midrule
        DB-RNNT+DB-LM \cite{Le2021ContextualizedSE-DB} 
        & \makecell{1.98 \\ (1.5/5.7)} 
        & \makecell{5.86 \\ (4.9/14.1)} 
        & \makecell{2.09 \\ (1.6/6.2)} 
        & \makecell{6.09 \\ (5.1/15.1)} 
        & \makecell{2.14 \\ (1.6/6.7)} 
        & \makecell{6.35 \\ (5.1/17.2)} \\
        \midrule
        \makecell[l]{Attention-based DB \\ + BPB beam search \cite{sudo2024contextualized}} 
        & \makecell{2.75 \\ (2.3/6.0)} 
        & \makecell{5.60 \\ (4.9/12.0)} 
        & \makecell{3.21 \\ (2.7/7.0)} 
        & \makecell{6.28 \\ (5.5/13.5)} 
        & \makecell{3.47 \\ (3.0/7.7)} 
        & \makecell{7.34 \\ (6.4/15.8)} \\
        \midrule
        \makecell[l]{Biasing fusion \\ + SpeechLLM \cite{gong24b_biasfusion+speechllm}} 
        & \makecell{1.6 \\ (1.3/5.5)} 
        & \makecell{3.8 \\ (2.6/13.5)} 
        & \makecell{1.9 \\ (1.4/6.0)} 
        & \makecell{3.9 \\ (2.7/14.2)} 
        & \multicolumn{2}{c}{--} \\
        \midrule
        \makecell[l]{CTC-Assisted LLM-Based ASR \\ \cite{yang2024ctc}} 
        & \makecell{1.27 \\ (1.00/3.67)} 
        & \makecell{2.72 \\ (2.16/8.02)} 
        & \makecell{1.33 \\ (1.03/3.92)} 
        & \makecell{3.04 \\ (2.40/9.04)} 
        & \makecell{1.33 \\ (1.00/4.16)} 
        & \makecell{2.99 \\ (2.31/9.33)} \\
        \midrule
        \SAPsq-TPI (Proposed) 
        & \makecell{1.12 \\ (0.89/\textbf{3.03})} 
        & \makecell{2.60 \\ (2.18/\textbf{6.55})} 
        & \makecell{1.16 \\ (0.93/\textbf{3.09})} 
        & \makecell{2.72 \\ (2.23/\textbf{7.29})} 
        & \makecell{1.22 \\ (0.96/\textbf{3.50})} 
        & \makecell{2.94 \\ (2.33/\textbf{8.62})} \\
        \bottomrule
    \end{tabular}
\end{table*}

\subsection{Experiment Results on SlideSpeech} \label{sec: slidespeech}

We first validate our method on SlideSpeech which contains natural long context OCR suitable for contextualized ASR scenarios. Table \ref{tab:main results} demonstrates the performance of the \SAPsq-TPI model. Under both 1-slide and 5-slide context settings, \SAPsq-TPI \textbf{achieves state-of-the-art} results on the two subsets of SlideSpeech. WER performance achieves a relative reduction of 30.19\% and 10.45\% over the previous SOTA MaLa-ASR on the S95 and L95 subsets, respectively. From the table, we can draw the follow conclusions:

\begin{itemize}
    \item  The SlideSpeech dataset features complex conference presentations regarding professional regions. The performance gap between uncontextualized and contextualized methods is obvious. Our \SAPsq-TPI using 5-slide context obtains a SOTA WER of 7.71\%
    \item \SAPsq-TPI demonstrates strong keywords integration abilities. B-WER of \SAPsq-TPI trained on L95 dataset decreases 41.1\% compared with Qwen2-Audio fine-tuned without contextual keywords. Notably, \SAPsq-TPI simultaneously maintains superior contextual keyword recall 95.59\%, demonstrating its effectiveness in both recognition accuracy and contextual relevance. B-WER/Recall performances of \SAPsq-TPI achieve an absolute improvement of 2.49\%/2.48\% and 0.35\%/0.28\% compared with MaLa-ASR on S95 and L95 subsets, respectively. 
    \item In the 5-slide scenarios compared to the 1-slide scenario, \SAPsq-TPI demonstrates stronger performance on the larger subset L95 while also improving overall WER and B-WER on the smaller subset S95. This reflects \SAPsq-TPI's robust capabilities in processing long-context information. We will further discuss robustness and generalization to longer contexts of \SAPsq-TPI in Section~\ref{sec: generalization}.
\end{itemize}

\subsection{Experiment Results on LibriSpeech} \label{sec: LibriSpeech}

To compare our \SAPsq-TPI method with more current contextualized ASR models, we conduct experiments on the widely used LibriSpeech dataset. 
We analyze WER(U-WER/B-WER) with biasing list lengths of 100, 500, and 1000 on test-clean and test-other subsets. The conclusions drawn from Table~\ref{tab:LibriSpeech comparision} align with those in Section \ref{sec: slidespeech}:

\begin{itemize}
    \item \SAPsq-TPI also achieves a \textbf{SOTA} performance on LibriSpeech across all biasing list settings, reaching an lowest WER of 1.12\%. \SAPsq-TPI achieves absolute reductions of 0.48\% and 0.74\% on WER compared with previous SOTA \citet{gong24b_biasfusion+speechllm} on test-clean (N=100) and test-clean (N=500), with even large absolute reductions of 0.8\% and 1.18\% on test-other (N=100) and test-other(N=100).
    \item \SAPsq-TPI exhibits strong robustness of processing longer context, increasing only 0.1\% WER on test-clean subset when 10x more biasing keywords are provided. B-WER on test-clean and test-other subsets also faintly increases when integrating more contextual keywords.
\end{itemize}

\subsection{Ablation Studies}

In this section, we aim to clarify the effectiveness of two-stage speech-aware context pruning and the proposed speech-driven attention-based pooling mechanism.

To investigate whether it's useful to prune long context in two stages, we compare our \SAPsq-TPI with directly concatenating keywords with speech features and jointly training pruning and recognition model.

\noindent\textbf{Prompt Concatenation (PC)} Following \citet{yang2024mala}, we directly concatenate contextual keywords with speech features. We design different instructions according to two situations: contextual keywords are provided, or no keywords are available. Please refer to the Appendix for detailed prompts.

\noindent\textbf{Joint Pruning-Integration (JPI)} Leveraging Qwen2-Audio’s multi-task learning capability, we explore training a single model for context pruning and speech recognition. We train the model to sequentially output pruned keywords and recognition results. Please refer to the Appendix for reasons and deductions.

Furthermore, to clarify the effectiveness of speech-driven attention-based pooling, we compare the ASR performance and time consumption of our proposed \SAPsq methods to Qwen2-Audio with different stage-wise styles. 

\begin{table}[ht]
  \caption{Performance comparison of \SAPsq-PC, \SAPsq-JPI and \SAPsq-TPI evaluated on SlideSpeech test dataset, trained on S95/L95 datasets.}
  \label{tab:methods comparison}
  \centering
  \resizebox{\linewidth}{!}{
  \begin{tabular}{llccccc}
    \toprule
    \multirow{2}{*}{\textbf{Train}} & 
    \multirow{2}{*}{\textbf{Model}} & 
     \multirow{2}{*}{\makecell{ \textbf{Contextual}\\ \textbf{Keywords}}}  & \multicolumn{4}{c}{\textbf{Test}} \\
    \cmidrule(lr){4-7} 
    & & &  \textbf{WER} & \textbf{U-WER} & \textbf{B-WER} & \textbf{Recall $\uparrow$}  \\
    \midrule
    \multirow{6}{*}{\makecell{S95\\(161h)}}
     & \SAPsq-PC  & 1slide&  \textbf{7.85} & \textbf{7.96} & 6.36 & 93.77 \\
     & \SAPsq-JPI  & 1slide&  8.12 &8.25  &6.39  &93.71  \\
     & \SAPsq-TPI  & 1slide  & 8.04  & 8.13 & \textbf{5.29} & \textbf{94.85} \\
     \cmidrule(lr){2-7}
     & \SAPsq-PC  & 5slides & 7.93 & 8.13 & 6.24 & 93.88 \\
     & \SAPsq-JPI  & 5slides  & 7.92 & \textbf{7.70} & 9.62 & 90.79 \\
     & \SAPsq-TPI  & 5slides&  \textbf{7.86} & 8.17 & \textbf{5.18} & \textbf{94.98} \\
    \midrule
    \multirow{6}{*}{\makecell{L95\\(473h)}} & \SAPsq-PC  & 1slide&  7.93 & \textbf{7.94} & 7.74 & 92.26 \\
     & \SAPsq-JPI  & 1slide&  8.07 & 8.08 & 7.85 & 92.31 \\
     & \SAPsq-TPI  & 1slide&  \textbf{7.92} & 8.16 & \textbf{4.71} & \textbf{95.47} \\
     \cmidrule(lr){2-7}
     & \SAPsq-PC  & 5slides& 7.86 &8.04 & 6.26 & 93.88 \\
     & \SAPsq-JPI  & 5slides &7.98  &8.05  &7.34  & 92.80 \\
     & \SAPsq-TPI  & 5slides& \textbf{7.71} & 8.08 & \textbf{4.54} & \textbf{95.59} \\
    \bottomrule
  \end{tabular}
  }
\end{table}

\begin{table}[h]
    \centering
    \caption{F1-score of filtered contextual keywords of \SAPsq-JPI and \SAPsq-TPI on SlideSpeech test dataset, trained on L95.}
    \begin{tabular}{lcccc}
        \toprule
        \multirow{2}{*}{\textbf{Model}} &  \multirow{2}{*}{\makecell{ \textbf{Contextual}\\ \textbf{Keywords}}} &\multirow{2}{*}{\textbf{F1-score}} \\ 
        & &  \\
        \midrule
        \SAPsq-JPI & 1slide  &  20.45  \\ 
        \SAPsq-TPI & 1slide & 93.85  \\ 
        \midrule
        \SAPsq-JPI & 5slides &  21.19 \\ 
        \SAPsq-TPI & 5slides &  94.48 \\
        \bottomrule
    \end{tabular}
    \label{tab:f1-score}
\end{table}

\subsubsection{Ablation Study on Speech-Aware Context Pruning}

In Table \ref{tab:methods comparison}, among the three proposed methods, \SAPsq-TPI demonstrates consistent superiority in biased keyword prediction accuracy, achieving the lowest B-WER across all configurations. Notably, \SAPsq-PC attains the optimal U-WER of 7.94\% on the L95 dataset, while its contextual precision lags behind \SAPsq-TPI by 2.03-3.15\% in B-WER metrics. \SAPsq-JPI exhibits overall WER degradation and reduced contextual sensitivity. Table \ref{tab:f1-score} demonstrates that \SAPsq-JPI achieves a lower F1-score of predicting speech-relevant contextual keywords, likely due to interference between its dual objectives of keyword selection and speech recognition. This performance gap highlights the critical importance of effective context pruning, where suboptimal keyword selection directly impacts recognition quality.

We can also conclude from Table \ref{tab:methods comparison} how the proposed three methods perform when contextual keywords are augmented. While \SAPsq-PC and \SAPsq-JPI show marginal improvements or even alter for the worse with extended context windows, \SAPsq-TPI achieves constant WER reductions of 2.2\% (S95) and 2.7\% (L95) under extended context conditions. This indicates that \SAPsq-TPI possesses superior contextual robustness, capable of effectively integrating long-form contextual information without performance saturation.

\subsubsection{Ablation Study on Speech-driven Attention-based Pooling}

\begin{table}[ht]
  \caption{Ablation study on effectiveness of Speech-driven attention-based pooling.}
  \label{tab:ablation wer}
  \centering
  \resizebox{\linewidth}{!}{
  \begin{tabular}{lclcccc}
    \toprule
    \multirow{2}{*}{\textbf{Train}} &  \multirow{2}{*}{\makecell{ \textbf{Contextual}\\ \textbf{Keywords}}} & \multirow{2}{*}{\textbf{Model}} & \multicolumn{4}{c}{\textbf{Test}} \\
    \cmidrule(lr){4-7} 
     &  & & \textbf{WER} & \textbf{U-WER} & \textbf{B-WER} & \textbf{Recall $\uparrow$}  \\
    \midrule
    \multirow{16}{*}{\makecell{L95\\(473h)}} & \multirow{8}{*}{1slide} & \textit{Qwen2-Audio} & & & \\
     & & +PC   & 8.22 & 8.35  & 6.34 & 93.82 \\
     & & +JPI  & 7.97 & 8.13 & 5.81 & 94.39 \\
     & & +TPI   & \textbf{7.92} & 8.19 & \textbf{4.12} & \textbf{96.00} \\
     \cmidrule(lr){3-7}
     & & \textit{\SAPsq} & & & \\
     & & +PC   & 7.93 &\textbf{7.94} & 7.74 & 92.26  \\
     & & +JPI  & 8.07  &8.05  &7.34  & 92.80  \\
     & & +TPI  & \textbf{7.92} & 8.16 & 4.71 & 95.47  \\
     \cmidrule(lr){2-7}
     & \multirow{8}{*}{5slides}  & \textit{Qwen2-Audio} & & & \\
     & & +PC   & 7.94 & 8.14  & 6.18 & 93.94 \\
     & & +JPI  & 7.99 & 8.19 & 5.26 & 94.94 \\
     & & +TPI  & 7.90 & 8.16 & \textbf{4.35} & \textbf{95.81} \\
     \cmidrule(lr){3-7}
     & & \textit{\SAPsq} & & & \\
     & & +PC   & 7.86 & \textbf{8.04} & 6.26 & 93.88  \\
     & & +JPI  & 7.98  &8.05  &7.34  & 92.80  \\
     & & +TPI  & \textbf{7.71} & 8.08 & 4.54 & 95.59  \\
    \bottomrule
  \end{tabular}
}
\end{table}

To clarify the effectiveness of Speech-driven attention-based pooling, we compare Qwen2-Audio models fine-tuned with the pooling strategy with those without. The experiments are conducted on SlideSpeech L95 dataset with 1-slide and 5-slide contextual keywords, evaluated on test set only. Table \ref{tab:ablation wer} illustrates that performances of PC and TPI improve when applied with Speech-driven attention-based pooling, while JPI deteriorates. The performance degradation of JPI may be because the pooling of text embeddings increases the ambiguity of context representations and intensifies the complexity of dual objectives. What's more, when larger amounts of contextual keywords are provided, the improvement brought by Speech-driven attention-based pooling becomes more obvious.

\begin{table}[h]
    \centering
    \caption{Comparison of TPI methods' training time of each stage with or without Speech-driven attention-based pooling.  Qwen2-Audio-TPI and \SAPsq-TPI are trained on L95 dataset (5 slides) with 7 40G A100 GPUs. ASR results are inferred on SlideSpeech 15-slide test set using 1 40G A100 GPU. }
    \renewcommand{\arraystretch}{1.5}
    \begin{tabular}{l cc}
        \toprule
        Method & \makecell{Training \\ time~(s)} & \makecell{Inference time~(s) \\ (15 slides)}\\ 
        \midrule
        Qwen2-Audio-PC & 60134.5 & 6054 \\
        \midrule
        \makecell{Qwen2-Audio-TPI \\(Stage \uppercase\expandafter{\romannumeral1} )} & 51003.8   & 1960 \\
        \makecell{Qwen2-Audio-TPI \\(Stage \uppercase\expandafter{\romannumeral2} )} & 30120.3  & 5047 \\
        \midrule
        \SAPsq-TPI (Stage \uppercase\expandafter{\romannumeral1} ) & 38267.9  & 1564 \\
        \SAPsq-TPI (Stage \uppercase\expandafter{\romannumeral2} ) & 29608.8  & 5076 \\
        \bottomrule
    \end{tabular}
    \label{tab:ablation time}
\end{table}

What's more, we analyze the training and inference time of Qwen2-Audio-PC, Qwen2-Audio-TPI and \SAPsq-TPI. We train three models on SlideSpeech L95 5-slide dataset using 7 40G A100 GPUs and infer on 15-slide dataset using 1 40G A100 GPU in Table \ref{tab:ablation time}. The training and inference time of \SAPsq-TPI is close to those of Qwen2-Audio-PC, indicating that two-stage pipeline doesn't introduce much complexity. For Qwen2-Audio-TPI and \SAPsq-TPI, the decrease in training time and inference time of Stage \uppercase\expandafter{\romannumeral2} is not obvious, as the number of contextual keywords is mostly less than 10 after context pruning. However, training time for context pruning significantly decreases by 24.97\%, inference time also decreases by 20.20\%. Our ablation study indicates that Speech-driven attention-based pooling plays an important role when there are large amounts of contextual keywords.

\subsection{Analysis on Pooling Window Size} \label{sec:pooling window}

\begin{figure}[h]
  \centering
  \includegraphics[width=\linewidth]{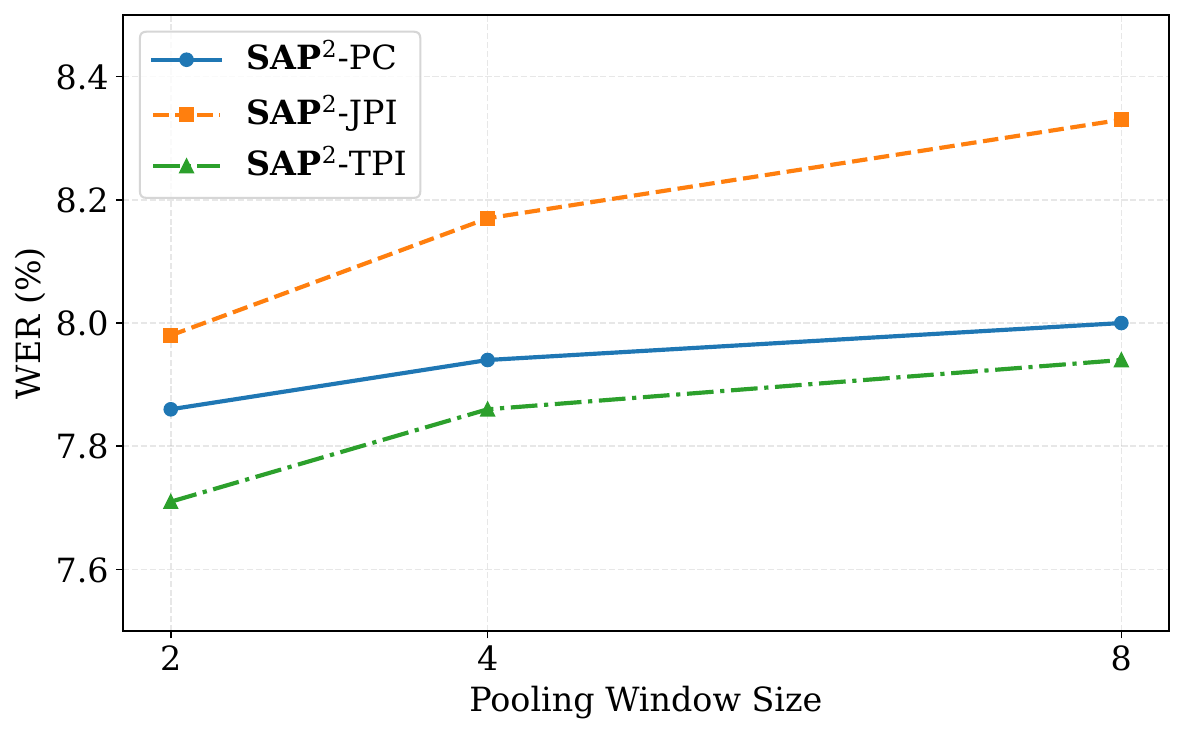}
  \caption{Analysis of the impact of pooling window sizes on WER. Experiments are conducted on L95 dataset (5 slides).}
  \label{fig: window size}
\end{figure}

To analyze model performance as the compression rate increases, we train \SAPsq-PC, \SAPsq-JPI and \SAPsq-TPI with pooling window sizes of 2, 4 and 8 on L95 (5 slides) dataset. Figure \ref{fig: window size} demonstrates that WER slightly increases when window size increases. However, \SAPsq-TPI constantly outperforms the other two methods, while \SAPsq-PC is most robust to pooling window size changes.

\subsection{Analysis of Inference Generalization Across Variable Context Lengths} \label{sec: generalization}

In Section~\ref{sec: slidespeech}, we analyzed the robustness of \SAPsq-TPI when trained and tested with varying numbers of slides. Since retraining the model for each context length is impractical, 
we specifically investigate whether \SAPsq-TPI exhibits stronger robustness than Qwen2-Audio-TPI on the challenging SlideSpeech dataset.

\begin{figure}[h]
  \centering
  \includegraphics[width=\linewidth]{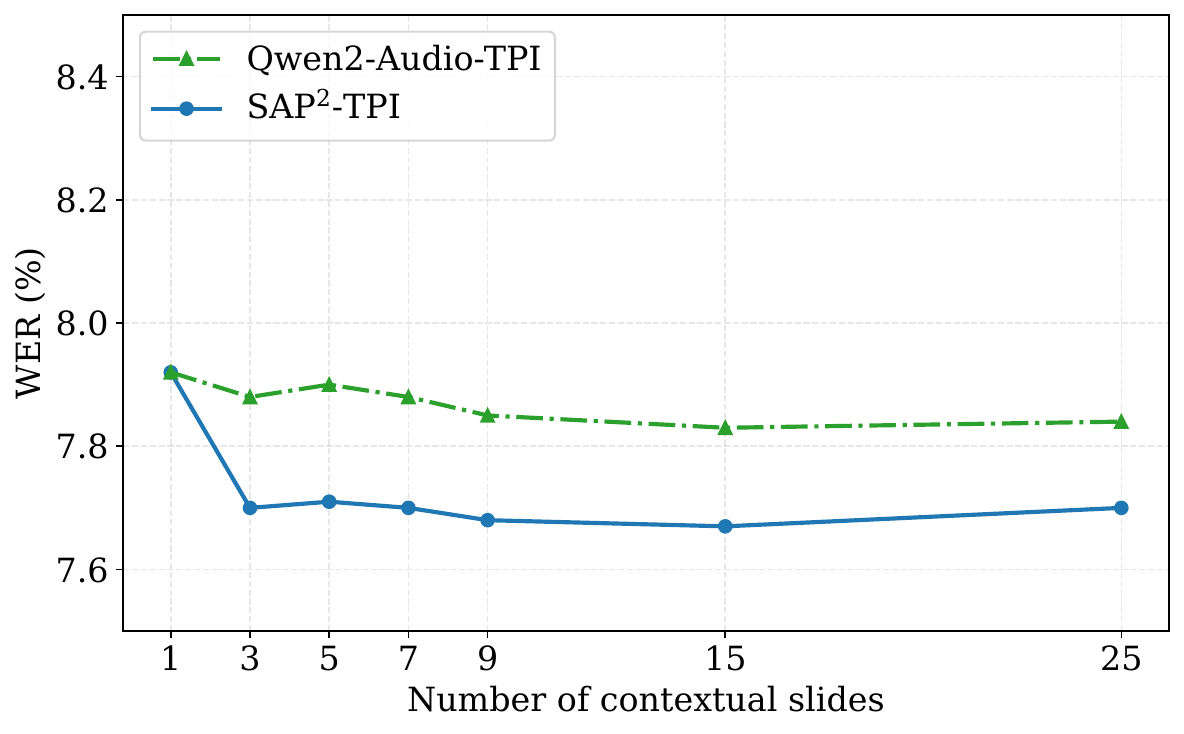}
  \caption{ WER across context lengths (1, 3, 5, 7, 9, 15, 25 slides) for \SAPsq-TPI and Qwen2-Audio-TPI, both trained on 5-slide segments (L95 dataset)}
  \label{fig: slides generalization}
\end{figure}

Both models (trained on L95 5-slide dataset) are evaluated on test sets with 1, 3, 5, 7, 9, 15, and 25 slides (Please refer to the Appendix for statistics of test sets). As shown in Figure~\ref{fig: slides generalization}, \SAPsq-TPI consistently outperforms the baseline when contextual slides exceed 3. Increasing slides from 5 to 15 expands the mean/median token lengths from 402.81/332 to 1024.88/964. The Keyword Coverage Rate rises from 6.51\% to 8.50\%, while the Information Rate drops from 0.64\% to 0.33\%. Despite these changes, WER slightly improves from 7.71\% to 7.67\%, demonstrating robust adaptation to longer contexts.

These results underscore \SAPsq-TPI’s superior generalization capability, with performance gains becoming increasingly significant as context length grows.

\section{Conclusion}

In this paper, we tend to address two limitations of contextualized ASR systems: (1) constrained context window capacity for document-level long-form processing, and (2) the performance degradation caused by noisy long context information. 
We propose \SAPsq~ that synergizes dynamic keyword pruning with speech-adaptive attention pooling for contextual embedding compression. Extensive experiments on SlideSpeech and LibriSpeech verify \SAPsq’s superiority
and stable performance under various context settings. 
Because of the limitations of computational resources and data, we only use contexts splitted as single words, future directions will explore semantically meaningful phrasal contexts and multimodal fusion with visual embeddings.  We also hope this study motivates future investigations into context-aware speech recognition paradigms.

\section*{Acknowledgements}

This work is supported by the National Key R\&D Program of China (No.2022ZD0116405).

\bibliography{aaai2026}

\clearpage

\setcounter{secnumdepth}{2} 

\appendix

\section*{Appendix}

\section{Related Works}

\subsection{Pre-trained Speech Models}
The field of speech recognition has made remarkable strides with unsupervised pre-training techniques like Wav2Vec 2.0 ~\citep{baevski2020wav2vec} and w2v-BERT~\citep{chung2021w2v}.
A key advancement in this area is Whisper~\cite{radford2023robust}, which is pre-trained in a weakly supervised manner on large-scale speech datasets. Its robust pre-training capabilities, coupled with a flexible Encoder-Decoder architecture, have made it a popular choice for a variety of downstream tasks~\cite{wang2024can, wang2024cieasr, attia2024kid}.
The emergence of large multimodal models has driven further progress, particularly for tasks that require the alignment of speech and language. Models like Seed-ASR~\citep{bai2024seed}, Qwen2-Audio~\citep{chu2024qwen2}, and FireRedASR~\citep{xu2025fireredasr} offer greater flexibility and precision across diverse speech-related tasks. 

\subsection{Context Compression}
Recent approaches to context compression in large language models (LLMs) focus on improving the efficiency of handling long contexts while retaining important information. 
\citet{chevalier2023adapting} compresses long contexts into summary vectors, which are then used as soft prompts to improve model performance without significantly increasing the computational cost. 
\citet{ge2023context} utilizes a pre-trained autoencoder to generate compact memory slots for context.
\citet{li2023compressing} proposes a method that prunes redundant information in the input context, effectively reducing its size while maintaining key contextual elements. 
LongLLMLingua~\citep{jiang2023llmlingua, jiang2023longllmlingua} focuses on prompt compression for long-context scenarios, using a mechanism to enhance the LLM's ability to retain key information in the compressed prompt. 
\citet{cao2024retaining} uses queries to guide the compression process, ensuring important contextual information is retained during compression even with high compression ratios. 

\subsection{Contextualized ASR}
Recent advances in Contextualized ASR have shown significant improvements by incorporating contextual information.
A common approach is deep biasing end-to-end models.
\citet{huang2023contextualized} utilizes textual context phrases to assist in training ASR models, improving performance in recognizing longer, context-dependent sequences.
\citet{sudo2024contextualized} uses an editable bias phrase list to inject domain-specific knowledge into the ASR system. 
\citet{yang2024promptasr} incorporates text prompts into end-to-end ASR systems, enabling flexible transcription control and contextual understanding. \citet{jayanthi2023retrieve} explores a two-stage retrieve-and-copy pipeline to retrieve related textual content based on similarity metrics.
The other approaches utilize SpeechLLM models for enhanced contextual modeling capabilities.
\citet{lakomkin2024end} uses LLaMA for speech recognition, treating the task as a mixed-modal language modeling problem and effectively incorporating textual contexts alongside audio inputs. 
\citet{yang2024mala} utilizes the WavLM Large model as a speech encoder, pre-trained on extensive audio data, and Vicuna as its LLM decoder.

\section{Derivation of Speech-driven Attention-based Pooling}

In SpeechLLM models, acoustic features $\boldsymbol{X}$ is transformed into $\boldsymbol{h}^x = {\{\boldsymbol{h}_i^x\}}_{i=1}^{T}  \in \mathbb{R}^{T \times d}$ by an MLP layer, contextual keywords $\boldsymbol{Z}$ is embedded into $\boldsymbol{h}^z = {\{\boldsymbol{h}_i^z\}}_{i=1}^{C} \in \mathbb{R}^{C \times d}$. Following the notation of multi-head cross-modal attention, we can compute the attention scores between $\boldsymbol{h}^x$ and $\boldsymbol{h}^z$ as
\begin{align}
    \boldsymbol{Q} &= \boldsymbol{h}^x \cdot \boldsymbol{W}_Q \in \mathbb{R}^{H \times T \times d/H}, \\
    \boldsymbol{K} &= \boldsymbol{h}^z \cdot \boldsymbol{W}_K \in \mathbb{R}^{H \times C \times d/H}, \\
    \boldsymbol{A} &= \text{Softmax}\left(\frac{\boldsymbol{Q} \boldsymbol{K}^\top}{\sqrt{d/H}} \right) \in \mathbb{R}^{H \times T \times C},
\end{align}
where $H$ is the number of attention heads. 

To derive per-context-token weights, we aggregate attention scores over the audio length dimension $T$:
\begin{equation}
    \boldsymbol{\alpha} = \frac{1}{T} \sum_{i=1}^T \boldsymbol{A}^{(i)} \in \mathbb{R}^{H \times C}.
\end{equation}

Suppose that a pooling module of window size $n$ reduces the length of the text embeddings from $C$ to $N_c = \frac{C}{n}$, then context embeddings can be grouped into $N_c$ windows:
\begin{align}
    \boldsymbol{h}^z &= [\boldsymbol{h}_{W_1}^z, \ldots, \boldsymbol{h}_{W_{N_c}}^z] \\
    &= [\boldsymbol{h}_{1:n}^z, \ldots, \boldsymbol{h}_{(j-1)\times n+1:j\times n}^z, \ldots, \boldsymbol{h}_{N_c -n+1:N_c}^z], \nonumber
\end{align}
where $W_j$ represents the $j$-th window. We expand the shape of $\boldsymbol{\alpha}$ to the hidden size of $\boldsymbol{h}^z$ and compute pooled context embeddings by weighting tokens within each window:
\begin{align}
    \alpha_{i, \boldsymbol{W}_j} &= \frac{\exp \alpha_i}{\sum_{i' \in \boldsymbol{W}_j}\exp \alpha_{i'}}, \\
    \hat{\boldsymbol{h}}_{\boldsymbol{W}_j}^z &= \sum\limits_{i \in \boldsymbol{W}_j} \alpha_{i, \boldsymbol{W}_j} \cdot \boldsymbol{h}_i^z, \\
    \hat{\boldsymbol{h}}^z &= [\hat{\boldsymbol{h}}_{\boldsymbol{W}_1}^z, \ldots, \hat{\boldsymbol{h}}_{\boldsymbol{W}_{N_c}}^z] \in \mathbb{R}^{N_c \times d}.
\end{align}

$\hat{\boldsymbol{h}}^z$ is the compressed embedding of contextual keywords that concentrate on the speech information.

\section{Experimental Setup}

\subsection{Dataset}

We conduct experiments on the \textbf{SlideSpeech}~\citep{wang2024slidespeech} dataset. This large-scale multimodal collection contains 1,705 YouTube-sourced conference videos, featuring synchronized 720p video streams, 16kHz audio recordings, and slide-derived OCR texts. The corpus uniquely integrates pre-processed slide content including extracted keywords and layout annotations. The training set is stratified into two subsets: L95 and S95, comprising 473 hours and 161 hours of audio data, respectively.

To investigate the impact of extended contextual information on speech recognition accuracy, and to evaluate the effectiveness of our proposed long-context pruning and integration framework, we augment SlideSpeech OCR contexts by extracting semantic keywords from five consecutive slides. Specifically, using the Jaccard index to group slides into coherent clusters, combining multiple consecutive slides together. 

To more fairly and comprehensively evaluate the performance of our proposed \SAPsq, we conduct experiments on the \textbf{LibriSpeech} \cite{LibriSpeech} corpus. For the training and validation sets, we follow the approach in \citet{Le2021ContextualizedSE-DB} to dynamically construct biasing lists. Specifically, for each sample, we first identify words in its ASR label that fall outside the common5k vocabulary. Then, we add $N$ (randomly selected between 400 and 800) distractors randomly sampled from the pool of 209.2k rare words. The test set remains consistent with \citet{Le2021ContextualizedSE-DB}. In line with most previous works, we evaluate our method under three settings: $N=\{100, 500, 1000\}$. We train separately on the 460-hour "train-clean" set and the 500-hour "train-other" set, and perform inference on the "test-clean" and "test-other" sets, respectively.

\subsection{SlideSpeech Test Set Configurations}

Here, we demonstrate statistics of contextual keywords for different test set settings in Table \ref{tab:slides}. As slide length increases, keyword coverage rate increases while information rate decreases, indicating that more relevant information is provided in more noisy contexts.

\begin{table*}[h]
    \centering
    \caption{Contextual keyword statistics corresponding to different slide length test sets}
    \begin{tabular}{c|c|c|c|c}
        \hline
        Slide Numbers & Average token length & Median token length & Keyword coverage rate & Information rate \\
        \hline
        1 slide & 64.57 & 53 & 6.66 & 2.56 \\
        3 slides & 311.57 & 280 & 9.04 & 3.03 \\
        5 slides & 402.81 & 322 & 10.08 & 2.22 \\
        7 slides & 588.64 & 549 & 10.78 & 1.76 \\
        9 slides & 713.41 & 692 & 11.38 & 1.53 \\
        15 slides & 1024.88 & 964 & 12.61 & 1.15 \\
        25 slides & 1562.73 & 1474 & 14.06 & 0.87 \\
        \hline
    \end{tabular}
    \label{tab:slides}
\end{table*}

\section{Clarification of Joint Pruning-Integration Ablation}

Relying on the multi-task ability and auto-regressive decoding of SpeechLLM models, to find out if the two-stage pipeline is necessary, we intend to train a single model to first prune contextual keywords, then recognize the speech according to pruned results. Following notations in the main paper, we can parameterize both stages through shared model parameters $\theta$. This allows us to reformulate the auto-regressive generation as:
\begin{align}
    &p_\theta (\boldsymbol{Y} | \boldsymbol{X}, \boldsymbol{Z}) = p_\theta (\boldsymbol{Y} | \boldsymbol{X}, \hat{\boldsymbol{Z}}) \cdot p_\theta (\tilde{\boldsymbol{Z}} | \boldsymbol{X}, \boldsymbol{Z}) \nonumber\\ 
    &= \prod_{j=1}^L p_{\theta}(\boldsymbol{y}_j|\boldsymbol{X}, \tilde{\boldsymbol{Z}}, \boldsymbol{Y}_{<j})\prod_{j=1}^{c}p_{\theta}(\tilde{\boldsymbol{z}}_j|\boldsymbol{X}, \boldsymbol{Z}, \tilde{\boldsymbol{Z}}_{<j}) \nonumber.
\end{align}

We can assume that when sequentially processing original and pruned keywords, the model's attention mechanism naturally focuses on pruned keywords for subsequent generation:
\[
p_{\theta}(\boldsymbol{y}_j|\boldsymbol{X}, \tilde{\boldsymbol{Z}}, \boldsymbol{Y}_{<j}) = p_{\theta}(\boldsymbol{y}_j|\boldsymbol{X}, \boldsymbol{Z}, 
 \tilde{\boldsymbol{Z}}, \boldsymbol{Y}_{<j}),
\]
leading to the unified generation formulation:
\[
p_\theta ([\tilde{\boldsymbol{Z}}, \boldsymbol{Y}] | \boldsymbol{X}, \boldsymbol{Z}) =\prod_{j=1}^{L+c}p_\theta(\tilde{\boldsymbol{y}}_j| \boldsymbol{X}, \boldsymbol{Z}, \tilde{\boldsymbol{Y}}_{<j}).
\]

This derivation demonstrates that training a single model to prune long context and recognize speech is also reasonable. We aim to investigate whether it's better to prune keywords explicitly in two stages or in this joint training approach.

\section{Instruction Prompts}

To instruct Qwen2-Audio to perform tasks of pruning or ASR, we design instruction prompts of TPI, PC and JPI under different conditions.

For speech-driven attention-based pooling, we only compress contextual keywords, excluding instruction prompts. To identify the positions of contextual keywords, we add special tokens \textit{<|startofcontext|>} and \textit{<|endofcontext|>}. Instruction templates do not contain special tokens when directly fine-tuning Qwen2-Audio.

\subsection{Prompt Concatenation}

The prompt template is dynamically selected based on keyword availability:
\begin{itemize}
\item \textit{“Transcribe speech to text according to keywords that may appear in the utterance. Possible keywords are: \{\}”} (with keywords)
\item \textit{“Transcribe speech to text.”} (without keywords)
\end{itemize}

\subsection{Two-stage Pruning-Integration}

The instruction of the context pruning model is \textit{"Select keywords that may appear in the speech from the following keywords list: \{\}"}. The instruction of the speech recognition model is the same as the \textbf{Prompt Concatenation} method.

\subsection{Joint Pruning-Integration}

The instruction template is: \textit{“First select keywords that may appear in the speech from given keywords list. Then transcribe speech to text according to selected keywords. Keywords are: \{\}”}, with responses formatted as \textit{“Selected keywords are: < keywords >. Transcription: < transcription >”}.

\section{More correction examples}

As illustrated in Figure 1 of our paper, we present two examples of recognizing rare personal names and medical terminology. These two examples demonstrate the effectiveness of our \SAPsq-TPI compared to MaLa-ASR and Qwen2-Audio-PC. Below, we will showcase more examples. In the Ground Truth, each \textit{\textbf{italicized}} text represents core keywords (included in the biasing list constructed from OCR texts). {\color{red}Red} texts indicate recognition errors in proper nouns, while {\color{green}green}-highlighted texts showcase corrections made by \SAPsq-TPI and other models.

\begin{figure}[!htbp]
    \centering
    \includegraphics[width=0.75\linewidth]{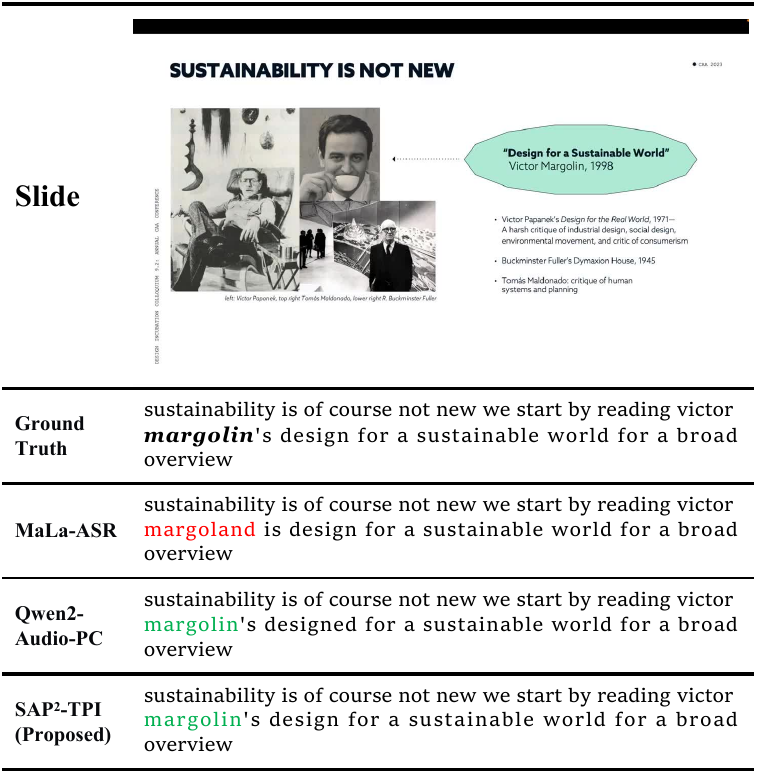}
    \caption{More correction examples in personal names.}
    \label{fig:person-name1}
\end{figure}

\begin{figure}[!htbp]
    \centering
    \includegraphics[width=0.75\linewidth]{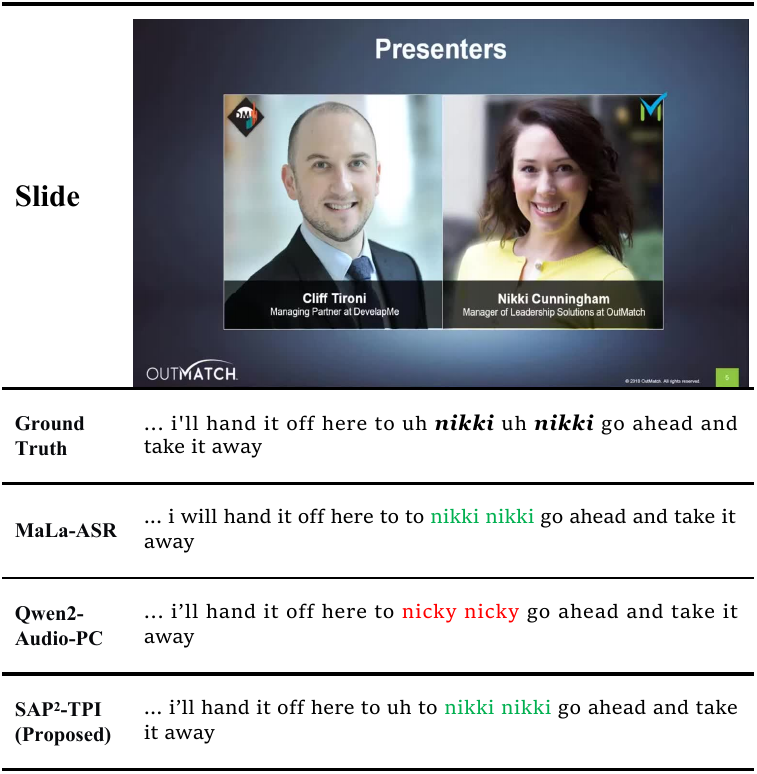}
    \caption{More correction examples in personal names.}
    \label{fig:person-name2}
\end{figure}

\begin{figure}[!htbp]
    \centering
    \includegraphics[width=0.75\linewidth]{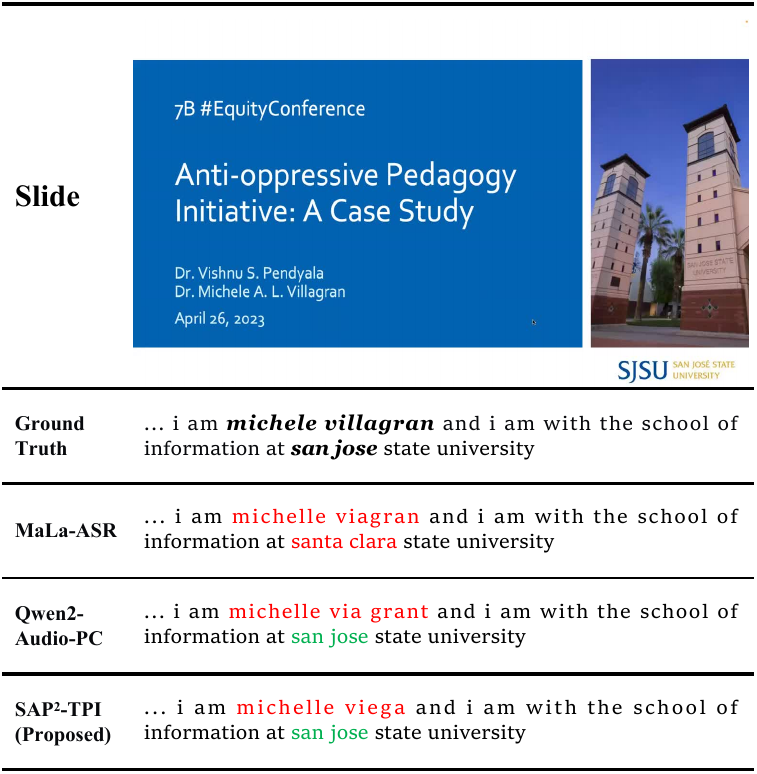}
    \caption{More correction examples in organization names.}
    \label{fig:3}
\end{figure}

\begin{figure}[!htbp]
    \centering
    \includegraphics[width=0.75\linewidth]{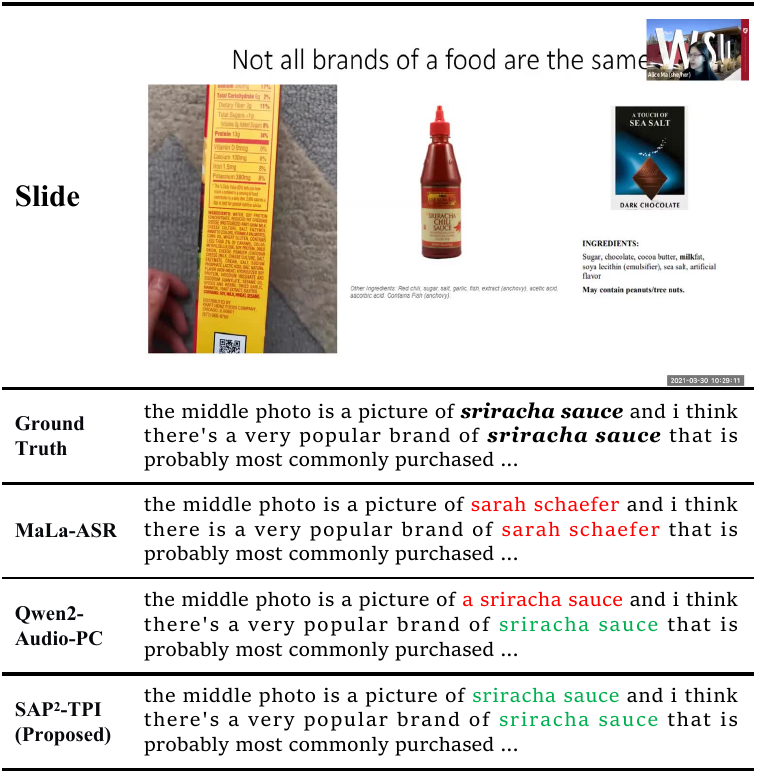}
    \caption{More correction examples in product names.}
    \label{fig:4}
\end{figure}

\begin{figure}[!htbp]
    \centering
    \includegraphics[width=0.75\linewidth]{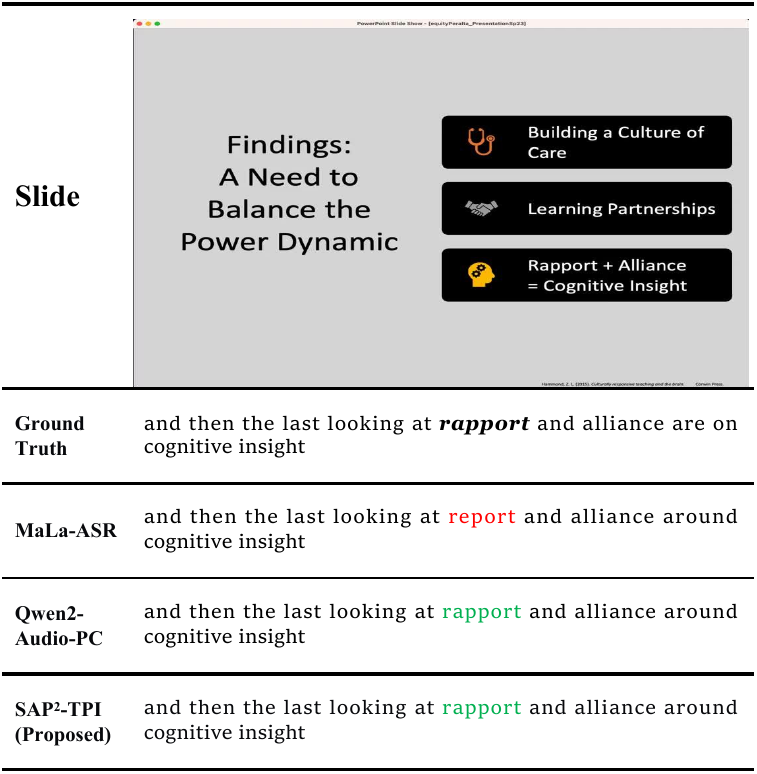}
    \caption{More correction examples in common nouns.}
    \label{fig:5}
\end{figure}

\newpage
\section{Training details}

We fine-tuned our \SAPsq-TPI model separately on four subsets:
\begin{itemize}
    \item The L95 (473 hours) and S95 (161 hours) subsets of the SlideSpeech dataset;
    \item The train-clean (460 hours) and train-other (500 hours) subsets of the LibriSpeech dataset, with artificially constructed biasing lists.
\end{itemize}

The model was then evaluated on the corresponding test sets for each subset. Table \ref{Table hyp} shows the detailed training hyper-parameter settings.

The model architecture of Qwen2-Audio comprises three key components: 1) a speech encoder for audio feature extraction, 2) a multimodal projector for modality alignment, and 3) a Qwen2-7B LLM backbone. During fine-tuning, we employ LoRA adapters (20.03M parameters) with rank 8 to update both the multimodal projector and LLM backbone while keeping the speech encoder frozen. 

We use the AdamW optimizer and cosine learning rate schedule which begins at 1e-4.

We utilize bfloat16 precision to balance computational efficiency and model performance fidelity while maintaining sufficient numerical stability.

We use 8 $\times$ NVIDIA A100-PCIE-40GB GPUs for training. Training on SlideSpeech S95 completes within 5 hours. Training on SlideSpeech L95, LibriSpeech test-clean, and test-other completes within 10 hours for each dataset separately.

\begin{table}[!htbp]
    \caption{Training hyper-parameters of the \SAPsq-TPI model in different datasets.}
    \label{Table hyp}
    \begin{tabular}{ccc}
    \toprule
    Hyper-parameters & SlideSpeech & LibriSpeech\\
    \midrule
    epochs & 1 & 1 \\
    train\_batch\_size & 2 & 2\\
    eval\_batch\_size & 2 & 2\\
    accum\_grad\_iters & 1 & 1\\
    lr\_scheduler\_type & cosine & cosine \\
    init\_lr & 1e-4 & 1e-4\\
    weight\_decay & 0.1 & 0.1\\
    lora\_rank & 8 & 32 \\
    lora\_alpha & 32 & 32 \\
    lora\_dropout & 0.05 & 0.05 \\
    Pooling Window Size & 2 & 2 \\
  \bottomrule
\end{tabular}
\end{table}

\end{document}